\documentclass[11pt]{article}
\usepackage{colacl}
\usepackage[dvips]{epsfig}
\author{ Eric Brill and Grace Ngai\\ Department of Computer Science\\ %
The Johns Hopkins University\\ Baltimore, MD 21218, USA\\ %
Email: {\tt \{brill,gyn\}@cs.jhu.edu}}
\title{\vspace{-65pt}{\normalsize \tt \hfill Appeared in \em{Proceedings of the 37th ACL}, 1999}\\ \mbox{} \\
Man\footnotemark[1] \space vs.\ Machine: A Case Study in Base %
Noun Phrase Learning} 

\begin{document}

\maketitle

\begin{abstract} 
A great deal of work has been done demonstrating the ability of
machine learning algorithms to automatically extract linguistic
knowledge from annotated corpora.  Very little work has gone into
quantifying the difference in ability at this task between a person
and a machine.  This paper is a first step in that direction.
\end{abstract}

\section{Introduction}
\renewcommand{\thefootnote}{\fnsymbol{footnote}}\
\footnotetext[1]{and Woman.}
\renewcommand{\thefootnote}{\arabic{footnote}}

Machine learning has been very successful at solving many problems in
the field of natural language processing.  It has been amply
demonstrated that a wide assortment of machine learning algorithms are
quite effective at extracting linguistic information from
manually annotated corpora.

Among the machine learning algorithms studied, rule based systems have
proven effective on many natural language processing tasks, including
part-of-speech tagging \cite{brill95:RBT,ramshaw94:tagging}, spelling
correction \cite{mangu97:cssc}, word-sense disambiguation
\cite{gale92:one_sense}, message understanding \cite{day97:alembic},
discourse tagging \cite{samuel98:discourse_tagging}, accent
restoration \cite{yarowsky94:decision_lists}, prepositional-phrase
attachment \cite{brill94:PPattach} and base noun phrase identification
\cite{ramshaw99:basenp,cardie98:basenp,veenstra98:basenp,argamon98:basenp}.
Many of these rule based systems learn a short list of simple rules
(typically on the order of 50-300) which are easily understood by
humans.

Since these rule-based systems achieve good performance while learning
a small list of simple rules, it raises the question of whether people
could also derive an effective rule list manually from an annotated
corpus.  In this paper we explore how quickly and effectively
relatively untrained people can extract linguistic generalities from a
corpus as compared to a machine.  There are a number of reasons for
doing this.  We would like to understand the relative strengths and
weaknesses of humans versus machines in hopes of marrying their
complementary strengths to create even more accurate systems.  Also,
since people can use their metaknowledge to generalize from a small
number of examples, it is possible that a person could derive
effective linguistic knowledge from a much smaller training corpus
than that needed by a machine.  A person could also potentially learn
more powerful representations than a machine, thereby achieving higher
accuracy.

In this paper we describe experiments we performed to ascertain how
well humans, given an annotated training set, can generate rules for
base noun phrase chunking.  Much previous work has been done on this
problem and many different methods have been used: Church's PARTS
\shortcite{church88:PARTS} program uses a Markov model; Bourigault
\shortcite{bourigault92:basenp} uses heuristics along with a grammar;
Voutilainen's NPTool \shortcite{voutilainen:NPTool} uses a lexicon
combined with a constraint grammar; Juteson and Katz
\shortcite{juteson95:basenp} use repeated phrases; Veenstra
\shortcite{veenstra98:basenp}, Argamon, Dagan \&
Krymolowski\shortcite{argamon98:basenp} and Daelemans, van den Bosch
\& Zavrel \shortcite{daelemans99:exceptions} use memory-based systems;
Ramshaw \& Marcus \shortcite{ramshaw99:basenp} and Cardie \& Pierce
\shortcite{cardie98:basenp} use rule-based systems.

\section{Learning Base Noun Phrases by Machine}

We used the base noun phrase system of Ramshaw and Marcus (R\&M) as
the machine learning system with which to compare the human learners.
It is difficult to compare different machine learning approaches to
base NP annotation, since different definitions of base NP are used in
many of the papers, but the R\&M system is the best of those that have
been tested on the Penn Treebank.\footnote{We would like to thank
Lance Ramshaw for providing us with the base-NP-annotated training and
test corpora that were used in the R\&M system, as well as the rules
learned by this system.}

To train their system, R\&M used a 200k-word chunk of the Penn
Treebank Parsed Wall Street Journal \cite{marcus93:penn_treebank}
tagged using a transformation-based tagger \cite{brill95:RBT} and
extracted base noun phrases from its parses by selecting noun phrases
that contained no nested noun phrases and further processing the data
with some heuristics (like treating the possessive marker as the first
word of a new base noun phrase) to flatten the recursive structure of
the parse.  They cast the problem as a transformation-based tagging
problem, where each word is to be labelled with a chunk structure tag
from the set \{I, O, B\}, where words marked ``I'' are inside some
base NP chunk, those marked ``O'' are not part of any base NP, and
those marked ``B'' denote the first word of a base NP which
immediately succeeds another base NP.  The training corpus is first
run through a part-of-speech tagger.  Then, as a baseline annotation,
each word is labelled with the most common chunk structure tag for its
part-of-speech tag.

After the baseline is achieved, transformation rules fitting a set of
rule templates are then learned to improve the ``tagging accuracy'' of
the training set.  These templates take into consideration the word,
part-of-speech tag and chunk structure tag of the current word and all
words within a window of 3 to either side of it.  Applying a rule to a
word changes the chunk structure tag of a word and in effect alters
the boundaries of the base NP chunks in the sentence.

An example of a rule learned by the R\&M system is: {\em change a
chunk structure tag of a word from I to B if the word is a determiner,
the next word is a noun, and the two previous words both have chunk
structure tags of I}.  In other words, a determiner in this context is
likely to begin a noun phrase.  The R\&M system learns a total of 500
rules.

\section{Manual Rule Acquisition}

R\&M framed the base NP annotation problem as a word tagging problem.
We chose instead to use regular expressions on words and part of
speech tags to characterize the NPs, as well as the context
surrounding the NPs, because this is both a more powerful
representational language and more intuitive to a person.  A person
can more easily consider potential phrases as a sequence of words and
tags, rather than looking at each individual word and deciding whether
it is part of a phrase or not.  The rule actions we allow
are:\footnote{The rule types we have chosen are similar to those used
by Vilain and Day \shortcite{vilain96:parsing} in transformation-based
parsing, but are more powerful.}
\begin{flushleft}
\begin{tabular}{lp{2.2in}}
{\bfseries A}dd & Add a base NP (bracket a sequence of words as a base
NP) \\  
{\bf K}ill & Delete a base NP (remove a pair of parentheses) \\
{\bf T}ransform & Transform a base NP (move one or both parentheses to
extend/contract a base NP) \\  
{\bf M}erge & Merge two base NPs
\end{tabular}
\end{flushleft}

As an example, we consider an actual rule from our experiments:
\begin{quote}
Bracket all sequences of words of: one determiner (DT), zero or more
adjectives (JJ, JJR, JJS), and one or more nouns (NN, NNP, NNS, NNPS),
if they are followed by a verb (VB, VBD, VBG, VBN, VBP, VBZ).
\end{quote}

In our language, the rule is written thus:\footnote{A full description
of the rule language can be found at
{\tt http://nlp.cs.jhu.edu/$\sim$baseNP/manual}.}

\begin{verbatim}
A
(* .)
({1} t=DT) (* t=JJ[RS]?) (+ t=NNP?S?) 
({1} t=VB[DGNPZ]?)
\end{verbatim}

The first line denotes the action, in this case, {\bf A}dd a
bracketing.  The second line defines the context preceding the
sequence we want to have bracketed \,---\, in this case, we do not
care what this sequence is.  The third line defines the sequence which
we want bracketed, and the last line defines the context following the
bracketed sequence.

Internally, the software then translates this rule into the more
unwieldy Perl regular expression:
\begin{small}
\begin{verbatim}
s{(([^\s_]+__DT\s+)([^\s_]+__JJ[RS]\s+)* 
([^\s_]+__NNP?S?\s+)+)([^\s_]+__VB[DGNPZ]\s+)}
{ ( $1 ) $5 }g
\end{verbatim}
\end{small}

The base NP annotation system created by the humans is essentially a
transformation-based system with hand-written rules.  The user
manually creates an ordered list of rules.  A rule list can be edited
by adding a rule at any position, deleting a rule, or modifying a
rule.  The user begins with an empty rule list.  Rules are derived by
studying the training corpus and NPs that the rules have not yet
bracketed, as well as NPs that the rules have incorrectly bracketed.
Whenever the rule list is edited, the efficacy of the changes can be
checked by running the new rule list on the training set and seeing
how the modified rule list compares to the unmodified list.  Based on
this feedback, the user decides whether to accept or reject the
changes that were made.  One nice property of transformation-based
learning is that in appending a rule to the end of a rule list, the
user need not be concerned about how that rule may interact with other
rules on the list.  This is much easier than writing a CFG, for
instance, where rules interact in a way that may not be readily
apparent to a human rule writer.

To make it easy for people to study the training set, word sequences
are presented in one of four colors indicating that they:

\begin{enumerate}
\item are not part of an NP either in the truth or in the output of the
person's rule set 
\item consist of an NP both in the truth and in the output of the
person's rule set (i.e. they constitute a base NP that the person's
rules correctly annotated)
\item consist of an NP in the truth but not in the output of the
person's rule set (i.e. they constitute  a recall error) 
\item consist of an NP  in the output of the person's rule set but not
in the truth (i.e. they constitute  a precision error)
\end{enumerate}

\begin{figure*}[htbp]
\begin{center}
\mbox{\epsfig{file=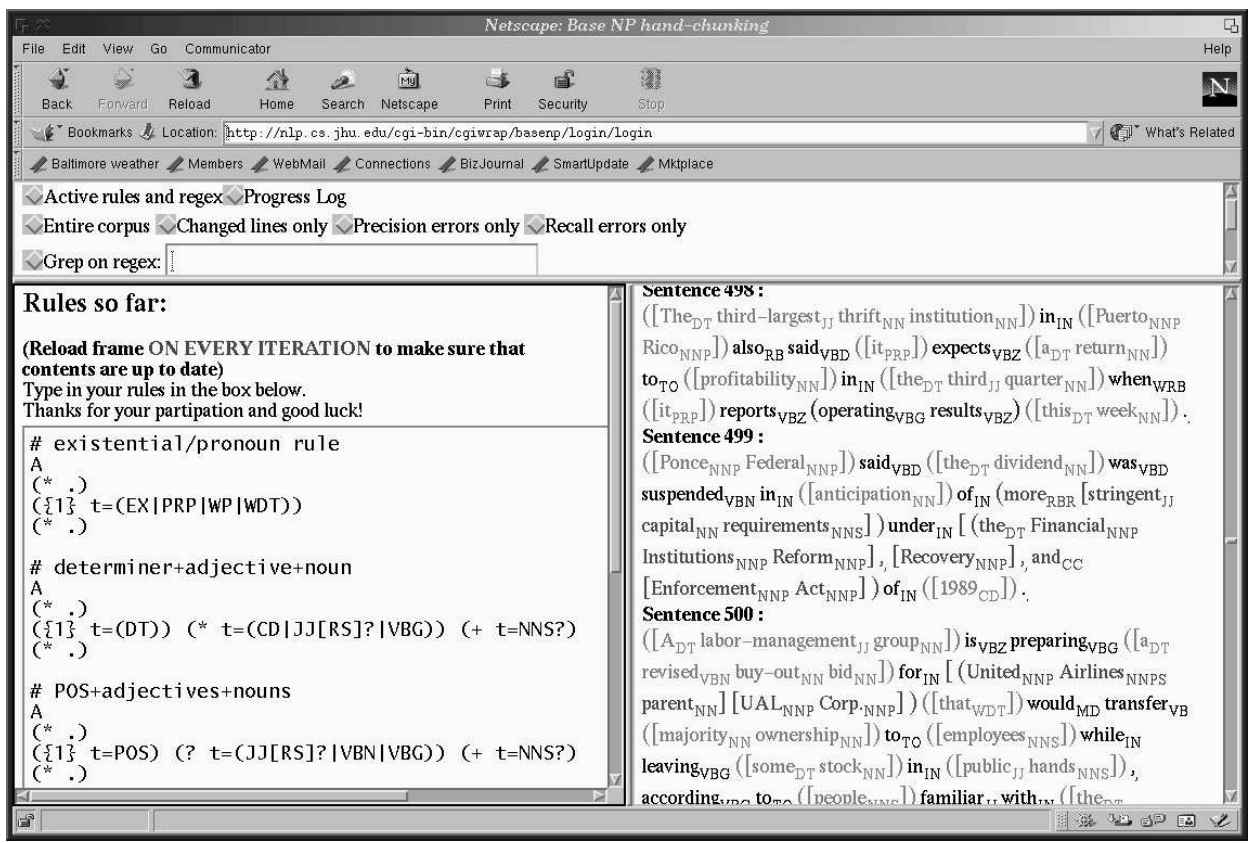, angle=-90, width=6.5in} }
\caption{ \label{fig:screenshot} Screenshot of base NP chunking system }
\end{center}
\end{figure*}

The actual system is located at \\ {\tt
http://nlp.cs.jhu.edu/$\sim$basenp/chunking}.  A screenshot of this
system is shown in figure \ref{fig:screenshot}.  The correct base NPs
are enclosed in parentheses and those annotated by the human's rules
in brackets.

\section{Experimental Set-Up and Results}

The experiment of writing rule lists for base NP annotation was
assigned as a homework set to a group of 11 undergraduate and graduate
students in an introductory natural language processing
course.\footnote{These 11 students were a subset of the entire class.
Students were given an option of participating in this experiment or
doing a much more challenging final project.  Thus, as a population,
they tended to be the less motivated students.}

The corpus that the students were given from which to derive and
validate rules is a 25k word subset of the R\&M training set,
approximately $\frac{1}{8}$ the size of the full R\&M training set.
The reason we used a downsized training set was that we believed
humans could generalize better from less data, and we thought that it
might be possible to meet or surpass R\&M's results with a much
smaller training set.

\begin{figure*}  
\begin{tabular}{|l|c|c|c|c||c|c|c|c|}
\hline
&\multicolumn{4}{|c||}{TRAINING SET (25K Words)}&\multicolumn{4}{|c|}{TEST SET}\\
\hline
 & Precision & Recall & F-Measure & $\frac{P+R}{2}$ & Precision & Recall &
 F-Measure & $\frac{P+R}{2}$ \\
\hline
Student 1  & 87.8\% & 88.6\% & 88.2 & 88.2 & 
	     88.0\% & 88.8\% & 88.4 & 88.4 \\ 
Student 2  & 88.1\% & 88.2\% & 88.2 & 88.2 & 
	     88.2\% & 87.9\% & 88.0 & 88.1  \\ 
Student 3  & 88.6\% & 87.6\% & 88.1 & 88.2 & 
	     88.3\% & 87.8\% & 88.0 & 88.1  \\ 
Student 4  & 88.0\% & 87.2\% & 87.6 & 87.6 & 
	     86.9\% & 85.9\% & 86.4 & 86.4 \\ 
Student 5  & 86.2\% & 86.8\% & 86.5 & 86.5 & 
	     85.8\% & 85.8\% & 85.8 & 85.8 \\ 
Student 6  & 86.0\% & 87.1\% & 86.6 & 86.6 & 
	     85.8\% & 87.1\% & 86.4 & 86.5 \\ 
Student 7  & 84.9\% & 86.7\% & 85.8 & 85.8 & 
	     85.3\% & 87.3\% & 86.3 & 86.3 \\ 
Student 8  & 83.6\% & 86.0\% & 84.8 & 84.8 & 
	     83.1\% & 85.7\% & 84.4 & 84.4 \\ 
Student 9  & 83.9\% & 85.0\% & 84.4 & 84.5 & 
	     83.5\% & 84.8\% & 84.1 & 84.2 \\ 
Student 10 & 82.8\% & 84.5\% & 83.6 & 83.7 & 
             83.3\% & 84.4\% & 83.8 & 83.8 \\ 
Student 11 & 84.8\% & 78.8\% & 81.7 & 81.8 & 
	     84.0\% & 77.4\% & 80.6 & 80.7 \\ 
\hline
\end{tabular}
\caption{\label{fig:results_students} P/R results of test subjects on
training and test corpora} 
\end{figure*}

Figure \ref{fig:results_students} shows the final precision, recall,
F-measure and precision+recall numbers on the training and test
corpora for the students.  There was very little difference in
performance on the training set compared to the test set. This
indicates that people, unlike machines, seem immune to overtraining.
The time the students spent on the problem ranged from less than 3
hours to almost 10 hours, with an average of about 6 hours.  While it
was certainly the case that the students with the worst results spent
the least amount of time on the problem, it was not true that those
with the best results spent the most time \,---\, indeed, the average
amount of time spent by the top three students was a little less than
the overall average \,---\, slightly over 5 hours.  On average, people
achieved 90\% of their final performance after half of the total time
they spent in rule writing.

The number of rules in the final rule lists also varied, from as
few as 16 rules to as many as 61 rules, with an average of 35.6
rules.  Again, the average number for the top three subjects was a
little under the average for everybody: 30.3 rules.

In the beginning, we believed that the students would be able to match
or better the R\&M system's results, which are shown in figure
\ref{fig:results_ramshaw}.  It can be seen that when the same training
corpus is used, the best students do achieve performances which are
close to the R\&M system's \,---\, on average, the top 3 students'
performances come within 0.5\% precision and 1.1\% recall of the
machine's.  In the following section, we will examine the output of
both the manual and automatic systems for differences.

\begin{figure*}
\begin{center}
\begin{tabular}{|c|c|c|c|c|}
\hline
Training set size(words)&Precision&Recall&F-Measure&$\frac{P+R}{2}$\\
\hline
25k  & 88.7\% & 89.3\% & 89.0  & 89.0 \\
200k & 91.8\% & 92.3\% & 92.0  & 92.1 \\
\hline
\end{tabular}
\end{center}
\caption{\label{fig:results_ramshaw} P/R results of the R\&M system
on test corpus} 
\end{figure*}

\section{Analysis}

Before we started the analysis of the test set, we hypothesized that
the manually derived systems would have more difficulty with potential
rules that are effective, but fix only a very small number of mistakes
in the training set.

The distribution of noun phrase types, identified by their part of
speech sequence, roughly obeys Zipf's Law \cite{Zipf35}: there is a
large tail of noun phrase types that occur very infrequently in the
corpus.  Assuming there is not a rule that can generalize across a
large number of these low-frequency noun phrases, the only way noun
phrases in the tail of the distribution can be learned is by learning
low-count rules: in other words, rules that will only positively
affect a small number of instances in the training corpus.

Van der Dosch and Daelemans \shortcite{daelemans98:full_mem} show that
not ignoring the low count instances is often crucial to performance
in machine learning systems for natural language.  Do the
human-written rules suffer from failing to learn these infrequent
phrases?

\begin{figure*}[htbp]
\begin{center}
\mbox{\epsfig{file=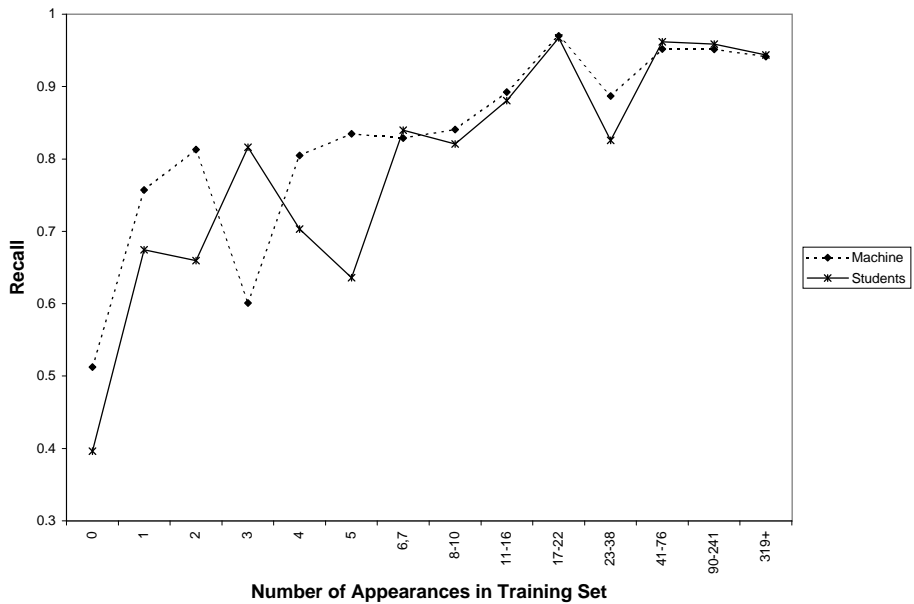, angle=-90, width=5.5in} }
\caption{ \label{fig:freqRcll} Test Set Recall vs.\ Frequency of
Appearances in Training Set. }
\end{center}
\end{figure*}

To explore the hypothesis that a primary difference between the
accuracy of human and machine is the machine's ability to capture the
low frequency noun phrases, we observed how the accuracy of noun
phrase annotation of both human and machine derived rules is affected
by the frequency of occurrence of the noun phrases in the training
corpus.  We reduced each base NP in the test set to its POS tag
sequence as assigned by the POS tagger. For each POS tag sequence, we
then counted the number of times it appeared in the training set and
the recall achieved on the test set.

The plot of the test set recall vs.\ the number of appearances in the
training set of each tag sequence for the machine and the mean of the
top 3 students is shown in figure \ref{fig:freqRcll}.  For instance,
for base NPs in the test set with tag sequences that appeared 5 times
in the training corpus, the students achieved an average recall of
63.6\% while the machine achieved a recall of 83.5\%.  For base NPs
with tag sequences that appear less than 6 times in the training set,
the machine outperforms the students by a recall of 62.8\% vs.\
54.8\%.  However, for the rest of the base NPs \,---\, those that
appear 6 or more times \,---\, the performances of the machine and
students are almost identical: 93.7\% for the machine vs.\ 93.5\% for
the 3 students, a difference that is not statistically significant.

The recall graph clearly shows that for the top 3 students,
performance is comparable to the machine's on all but the low
frequency constituents.  This can be explained by the human's
reluctance or inability to write a rule that will only capture a small
number of new base NPs in the training set.  Whereas a machine can
easily learn a few hundred rules, each of which makes a very small
improvement to accuracy, this is a tedious task for a person, and a
task which apparently none of our human subjects was willing or able
to take on.  

There is one anomalous point in figure \ref{fig:freqRcll}.  For base
NPs with POS tag sequences that appear 3 times in the training set,
there is a large decrease in recall for the machine, but a large
increase in recall for the students.  When we looked at the POS tag
sequences in question and their corresponding base NPs, we found that
this was caused by one single POS tag sequence \,---\, that of two
successive numbers (CD).  The test set happened to include many
sentences containing sequences of the type:
\begin{quote}
{\tt \ldots ( CD CD ) TO ( CD CD )\ldots }
\end{quote}
as in:
\begin{quote}
{\tt
( International/NNP Paper/NNP ) fell/VBD ( 1/CD $\frac{3}{8}$/CD ) to/TO
( 51/CD $\frac{1}{2}$/CD )\ldots
}
\end{quote}
while the training set had none.  The machine ended up bracketing
the entire sequence 
\begin{quote}
{\tt 1/CD $\frac{3}{8}$/CD to/TO 51/CD $\frac{1}{2}$/CD }
\end{quote}
as a base NP. None of the students, however, made
this mistake.

\section{Conclusions and Future Work}

In this paper we have described research we undertook in an attempt to
ascertain how people can perform compared to a machine at learning
linguistic information from an annotated corpus, and more importantly
to begin to explore the differences in learning behavior between human
and machine.  Although people did not match the performance of the
machine-learned annotator, it is interesting that these ``language
novices'', with almost no training, were able to come fairly close,
learning a small number of powerful rules in a short amount of time on
a small training set.  This challenges the claim that machine learning
offers portability advantages over manual rule writing, seeing that
relatively unmotivated people can near-match the best machine
performance on this task in so little time at a labor cost of
approximately US\$40.

We plan to take this work in a number of directions.  First, we will
further explore whether people can meet or beat the machine's accuracy
at this task.  We have identified one major weakness of human rule
writers: capturing information about low frequency events.  It is
possible that by providing the person with sufficiently powerful
corpus analysis tools to aide in rule writing, we could overcome this
problem.

We ran all of our human experiments on a fixed training corpus size.
It would be interesting to compare how human performance varies as a
function of training corpus size with how machine performance varies.

There are many ways to combine human corpus-based knowledge extraction
with machine learning.  One possibility would be to combine the human
and machine outputs.  Another would be to have the human start with
the output of the machine and then learn rules to correct the
machine's mistakes.  We could also have a hybrid system where the
person writes rules with the help of machine learning.  For instance,
the machine could propose a set of rules and the person could choose
the best one.  We hope that by further studying both human and machine
knowledge acquisition from corpora, we can devise learning strategies
that successfully combine the two approaches, and by doing so, further
improve our ability to extract useful linguistic information from
online resources.

\section*{Acknowledgements}

The authors would like to thank Ryan Brown, Mike Harmon, John
Henderson and David Yarowsky for their valuable feedback regarding
this work.  This work was partly funded by NSF grant IRI-9502312.

\bibliographystyle{acl}
\bibliography{references}
\end{document}